\documentclass{article}

\usepackage{PRIMEarxiv}

\usepackage[utf8]{inputenc} 
\usepackage[T1]{fontenc}    
\usepackage{url}            
\usepackage{booktabs}       
\usepackage{amsfonts}       
\usepackage{nicefrac}       
\usepackage{microtype}  
\usepackage{amsmath}
\usepackage{siunitx}
\usepackage[numbers]{natbib}
\usepackage{enumitem}
\usepackage{fancyhdr}       
\usepackage{graphicx}       
\graphicspath{{media/}}     

\title{Estimating productivity gains in digital automation
}

\author{
  Mauricio Jacobo-Romero$^1$,  Danilo S. Carvalho$^1$,  Andr\'e Freitas$^{1,2}$ \\
  ${}^1$Department of Computer Science, The University of Manchester, Manchester, UK\\
  \texttt{\{mauricio.jacoboromero, danilo.carvalho, andre.freitas\}@manchester.ac.uk} \\
  ${}^2$Reasoning \& Explainable AI group, Idiap Research Institute, Martigny, Switzerland \\
  \texttt{andre.freitas@idiap.ch}
}

\begin{document}
\maketitle

\begin{abstract}
This paper proposes a novel productivity estimation model to evaluate the effects of adopting Artificial Intelligence (AI) components in a production chain. Our model provides evidence to address the "AI's" Solow's Paradox. We provide (i) theoretical and empirical evidence to explain Solow's dichotomy; (ii) a data-driven model to estimate and asses productivity variations; (iii) a methodology underpinned on process mining datasets to determine the business process, BP, and productivity; (iv) a set of computer simulation parameters; (v) and empirical analysis on labour-distribution.  These provide data on why we consider AI Solow's paradox a consequence of metric mismeasurement. 
\end{abstract}


\section{Introduction}
Every time we integrate new technology into a company's operation, we assume it will increase the current productivity; because the same volume of production would be reached out  using smaller inputs of other factors of production,  or,  with the same resources, we might increase the production \cite{jalowiecki2020big}. Solow-Swan's model showed that increments in capital and labour only conduct to a time increase until the organisation reaches a steady-state growth. Meanwhile, technological changes deliver permanent growth \cite{Solow1957}. However, in the 1970s and 1980s, the computing revolution, the massive introduction of computers in all industry sectors, did not create significant productivity growth \cite{10.5555/171750.171780}. The Nobel-Prize-winning economist, Robert Solow, noticed this abnormality and stated: "we see the computer age everywhere except in the productivity statistics". This argument is called the "Solow's paradox" \cite{ByRobertM.Solow1987WBWO}.  

Since Solow's paradox formulation, it caught the researcher's attention. By the mid-1990s, they proposed four groups of explanatory arguments: (1) false hopes, (2)  productivity mismeasurement, (3) concentrated distribution and  rent dissipation, and (4) implementation and restructuring lags\cite{Brynjolfsson2017}. 

In 1996, researchers finally deciphered the paradox: “We conclude that the productivity paradox disappeared by 1991, at least in our sample of firms” \cite{Brynjolfsson1996}. According to this study, productivity models were based on loosely limited statistical analysis; these prototypes underestimated the economic benefits of computers. 
Additionally, new information systems implementation was problematic due to a shortage of specialised personnel.

Our central motivation is developing a methodology to interpret how digital automation (accelerated by recent advances in the AI space) impacts production chains and causes labour displacements. Relevant literature contains few models to forecast the effect of AI integration into business processes. These techniques require considerable amounts of interventional data to produce results. Unfortunately, well-controlled interventional studies in production chains are not  widely available\cite{Dongen2020}.

With these requirements and constraints in mind, we sought to: (i) provide theoretical and empirical evidence to address the Solow’s paradox (”lack of evidence of productivity improvement for AI-based interventions”); (ii) develop a micro-economic level, data-driven model to estimate and assess productivity variations after automation interventions within business process (BP) workflows; (iii) building a methodology that uses \textit{process mining} to support productivity estimation and analysis within BP workflows; (iv) determine a systematic method to parametrise productivity estimations for BP workflows; (v) and provide an empirical analysis of labour-redistribution of the model on a well-document public case study. 

This work is organised as follows: Section II provides related work, and Section III introduces key economic modelling concepts. Section IV describes our methodology. The proposed model is explained in section V. Section VI and VII introduce results and discussion, respectively. Lastly, we present our conclusions.

\section{Related Work}

In the mid-2010s, early Artificial Intelligence (AI) firm adopters stated that the acquisition of Natural language Processing (NLP) systems generated savings and productivity increments \cite{Bonsay_Cruz_Firozi_Camaro_2021}. However, empirical evidence showed that productivity benefits were not easy to track \cite{Denning2021}. AI integration exhibited the conditions of a Solow's Paradox redux.

This phenomenon motivated several works. They aimed to identify the reasons for the absence or emergence of AI productivity effects \cite{Schweikl2020}. 
These studies followed two main approaches: pessimistic reading of the empirical past and optimism about the future \cite{Brynjolfsson2017}. The first group suggests that AI provided benefits,  but statistics did not accurately capture them. The second one states that AI advantages are not yet part of the business operation \cite{63379290255c45f3b8ceb002d050138d}. Both perspectives employ macroeconomic information and lack the suitable tools to seize the data up to the task level \cite{Brynjolfsson2017}. As a consequence,  productivity assessment transfers AI benefits to other evaluation axes. The outcome, then, is an averaged result\cite{back2022return}. On the other hand, the optimistic course implies that there is a period in which technology is not mature enough to produce a discernible influence on productivity growth\cite{Acemoglu2014, jalowiecki2020big}.

It is clear, then, that a model capable of measuring productivity variations at a task level can provide the methodological support to disambiguate the current controversies. Some other works have found that a mismatch between technology and workers' skills negatively affects productivity \cite{Du2022}. AI systems assemble new "hybrid" duties that become part of the automated process\cite{Acemoglu2018}. The analyses conclude that labour composition is another factor that might harm productivity estimation \cite{Jackson2019, Li2022}. In other words, AI systems might require more labour to execute the automated BP \cite{Lu2021}. Hence, process inputs would change, and so the production volume \cite{Bonsay_Cruz_Firozi_Camaro_2021}. Therefore, labour composition is another relevant factor for AI productivity assessment.

Apart from these two main study segments, we identified a third one: AI Capital. AI capital is the portion of the income generated by those activities automated with AI systems \cite{Jones2010}. AI cash inflows are not challenging to track \cite{brynjolfsson2018artificial}. Unfortunately, it is complex to determine the root cause of these income streams \cite{Arnold2013}. Some authors state that market conditions are responsible for the generated wealth \cite{Acemoglu2018b}. In other words, AI Capital models require considerable data to produce results. 

The three approaches employ either Cobb-Douglas equations or variations to represent companies' production \cite{Acemoglu2018, Eggertsson2019, Schweikl2020}. This function family is exponential \cite{Solow1957}. Hence, parameter determination requires considerable effort \cite{Jackson2019}. To ease computations, researchers employ logarithmic techniques. In this fashion, linear regression methods are suitable to determine production parameters\cite{Acemoglu2018b}. Unfortunately, business processes store averaged information across different databases\cite{Arnold2013}. Hence, AI-systems contributions dilute across the business \cite{Acemoglu2018b}. Determining the productivity gains behind digital technologies provides a fundamental translation of value between the technology investment and the financial results \cite{jalowiecki2020big, Schweikl2020}.

To tackle this issue, we explored emerging process mining frameworks which induce and business processes based on event logs\cite{VanderAalst2012}, where an event is the basic unit of a BP. Each event provides basic information such as completion timestamp, executor, and activity costs\cite{Diba2020}. There are several process mining log formats. However, process mining researchers and practitioners adopted the IEEE XES format as a standard \cite{Garcia2019} which has a formal and canonical schema, recording compulsory fields and offering the possibility of including customised data in XML structure\cite{Munoz-Gama2022}. We found that process mining would provide adequate granularity and a feasibility argument to facilitate the measurement of productivity changes, dialoguing with the lower barriers provided by logs in contrast to competing more formal frameworks (such as Business Process management - BPM workflows).

Our examination of the relevant literature revealed a three-point gap: 1) labour productivity analysis should cover the task level characterisation\cite{Brynjolfsson2017,Jacobo-Romero2021}, 2) the interest in Solow's Paradox grew more towards measuring the impact of shifts in labour input than towards the identification of the sources of AI capital streams\cite{Macdonald2000}, 3) process mining might provide the required data to produce better productivity evaluations\cite{Munoz-Gama2022}.

\section{Estimating Productivity: Key Definitions}

In this section, we outline key critical definitions used throughout this paper. 

\noindent \textbf{Business Processes (BPs).} A Business Process (BP) is a series of coordinated activities that deliver a service or product\cite{Atrill2017}. These production chains are usually represented with Business Process Model and Notation(BPMN)\cite{Lima2012}. Each task is an abstraction of a production step that is parametrised in terms of production time and skill-set distribution \cite{Kossak2016}. These sketches provide information on task ownership, dependencies, information inputs and outputs \cite{Kalnins2004}.
    
\noindent \textbf{Productivity.} Productivity is the relation between the output generated by a BP and the required resources to create this result \cite{Prokopenko1987}. 
    Then, productivity can be described as follows: 
            \begin{equation}
                P = \frac{Y}{T}.
                \label{eq:eq1}
            \end{equation}
\noindent where production/output, $Y$, is the estimated volume of products/services generated by a firm, and $T$ is the labour input, the number of worked hours, to produce these outputs \cite{Macdonald2000}.

\noindent \textbf{Solow’s paradox.} "‘You can see computers everywhere but in the productivity statistics’, wrote Robert Solow in 1987. His dictum spawned several decades of economic research aimed at solving the mystery that has become known as the ‘Solow Paradox’: massive investment in computers but no net gain in productivity" \cite{Denning2021}. "Solow's paradox" attracted the attention of researchers and, after several years, some examinations concluded the paradox disappeared, at least for the analysed sample: "our sample consisted entirely of relatively large ‘Fortune 500’ firms" \cite{Brynjolfsson1996}. From the proposed conjectures, mismeasurement, was one of the most analysed. "The closer one examines the data behind the studies of IT performance, the more it looks like mismeasurement is at the core of the 'productivity paradox'" \cite{Brynjolfsson1996B}. 

Based on this, we decided to orchestrate our efforts toward expanding an answer for the "Solow's Paradox". Thus, we propose a model to analyse productivity variations due to the introduction of automation changes, in the digital context, motivated by the growth in the adoption of AI. 

\noindent \textbf{Stochastic queues.} A stochastic queue is a system that consists of two parts: a server and a queue. An event item arrives in the queue. The time for the item to be processed is dependent on the waiting (queueing) time and the service (processing) time\cite{Ross97}. Our strategy is to substitute each BP activity for a queue system. Each system belongs to a specific queue category. In this manner, it is possible to simulate the introduction of automation in a given BP. We defined three types of queue systems, dialoguing with the Cobb-Douglas equation: low-skilled workers, high-skilled workers and automated systems. Queue system type selection depends upon the activity description.
     
 \section{Methodology}
\begin{figure}[htbp]
            \centering 
                {\includegraphics[scale=0.4]{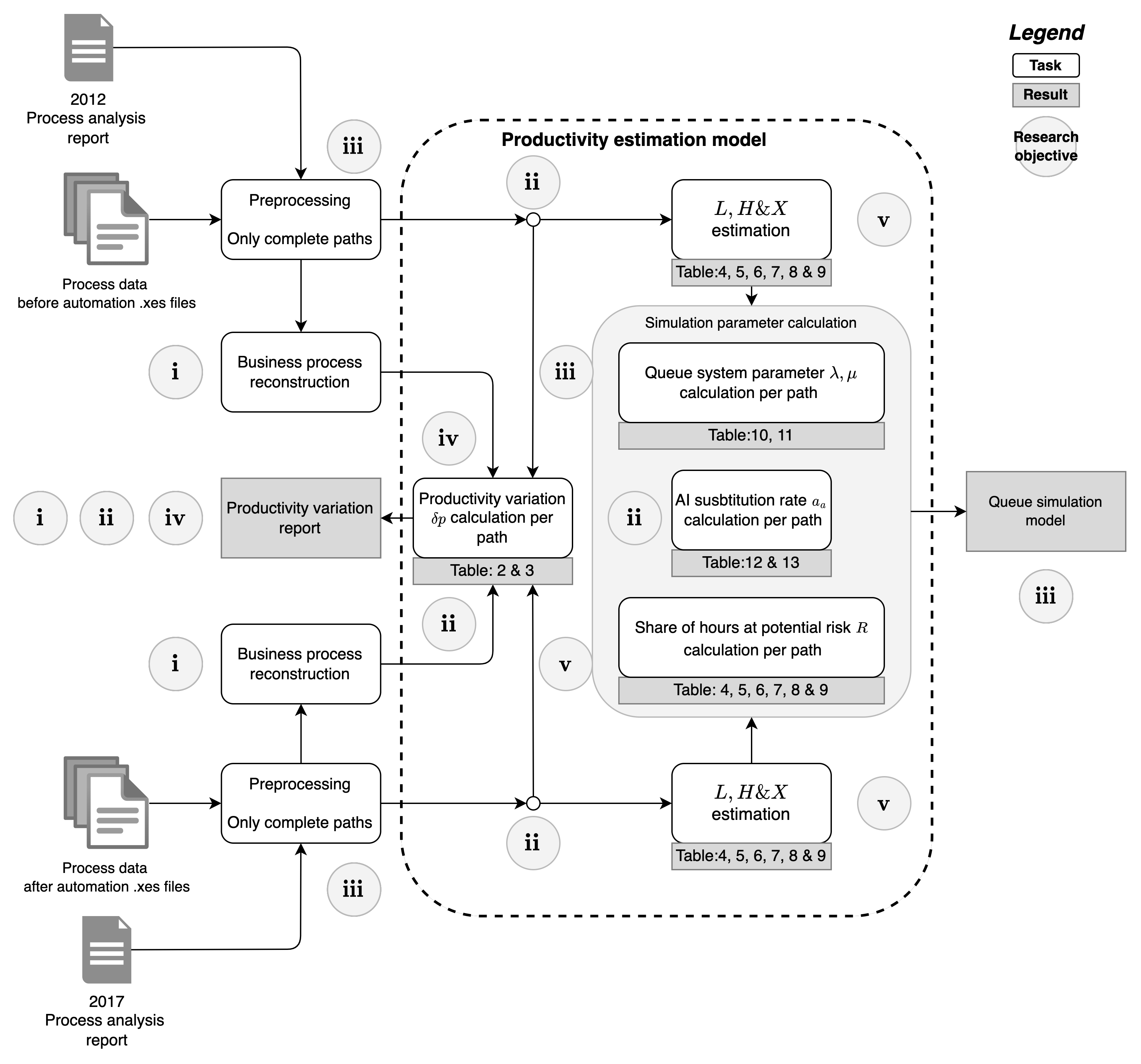}} 
            \caption{Experiment modelling diagram.}
            \label{fig:EXPSTUP} 
        \end{figure}

 We propose a novel method to estimate productivity and labour distribution changes of BPs after the introduction of automation. For that purpose, we designed the experiment described in figure \ref{fig:EXPSTUP}.

We analysed two datasets, 2012 and 2017, from the BPI Challenge. They describe a process that belongs to the same financial institution\cite{vanDongen2012, vanDongen2017}. 
The process changed over the time. Automation was integrated into the process. Thus, we found information about the BP state before and after the intervention.
    
As part of the BPI Challenge, participants produced a report on every workflow being represented via the logs. We employed this information to identify the labour composition of  the BP.

We applied the proposed parametrised queue-based productivity model and were able to produce a report on productivity fluctuations and queue system parameters. 
Our model presented a novel method to analyse productivity variations a priori and posteriori with minimum information requirements (building upon existing process mining frameworks). Moreover, the same model can be used to report labour displacements.

\section{Proposed model}

Economists usually express production functions as Cobb-Douglas equations. This family of functions are exponential and depends on two variables: \textit{labour} and \textit{cost of capital}. The exponent is known as the \textit{elasticity factor} and reflects the impact of these two inputs in the production\cite{Heathfield1971}.  
    
Unfortunately, the Cobb-Douglas functions do not provide information on labour distribution. Thus, we employed an equivalent production function:
        \begin{equation*}
                        Y = (L + AX)^{\alpha} H^{1 - \alpha}
        \end{equation*}

\noindent where $Y$ is the production of some good, that employs high-skilled labour hours, $H$, low-skilled labour hours, $L$, an input good that substitutes low-skilled labour, $X$, and applies $\alpha \in (0,1)$ to trace the relative shares of the high and low-skilled labour at a rate $A$\cite{Jackson2019}.

\begin{figure}
                        \centering
                            \begin{minipage}{.5\textwidth}
                                \centering
                                 \includegraphics[width=.95\linewidth]{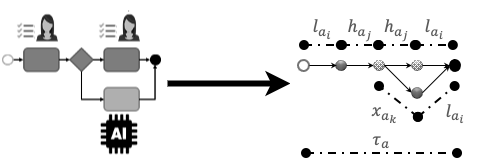}
                                 \caption{BP task mapping.}
                                 \label{fig:BPTaskMapping}
                            \end{minipage}%
                            \begin{minipage}{.5\textwidth}
                                \centering
                                 \includegraphics[width=.5\linewidth]{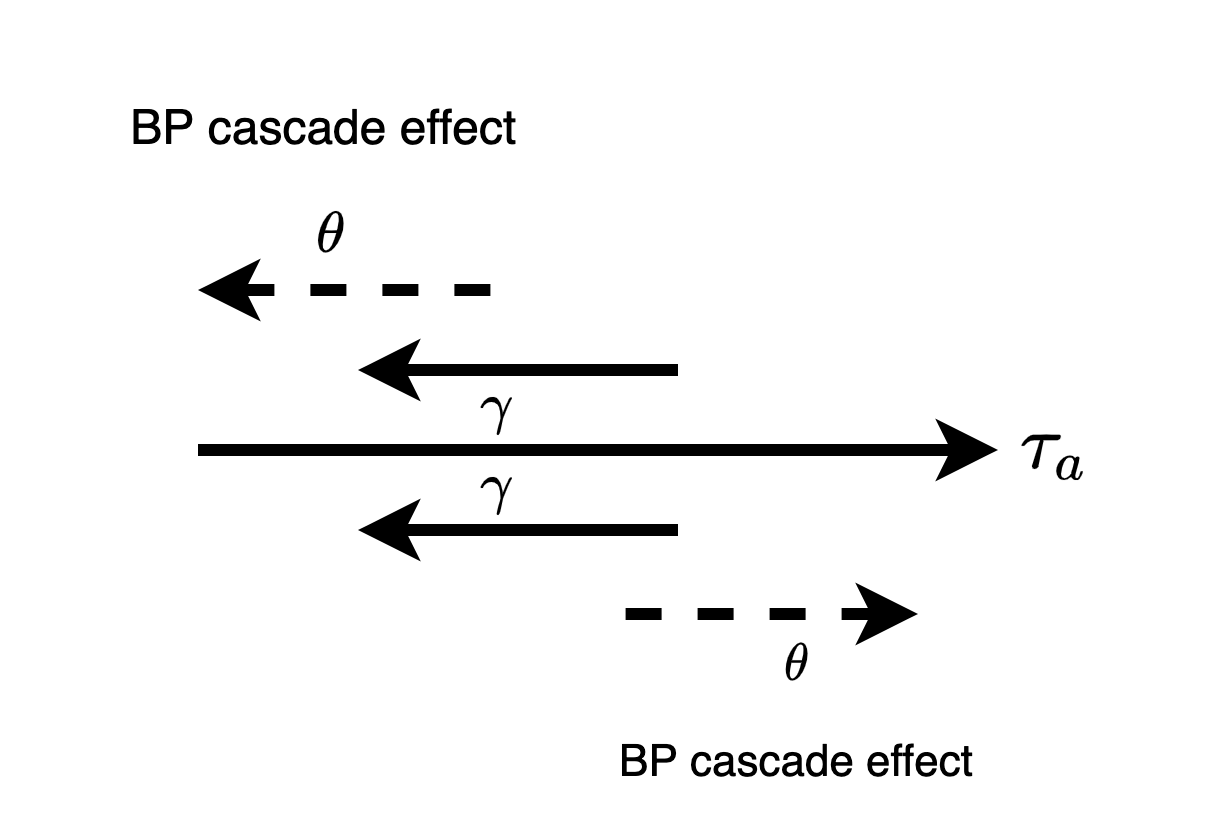}
                                 \caption{$\psi$ composition.}
                                 \label{fig:BPPhiVal}
                            \end{minipage}%
                    \end{figure}

Additionally, we integrated the following postulates:

    \begin{itemize}
        \item Production will keep constant after the introduction of a localised automation intervention. We assumed that no additional material/financial resources would be added to the BP.
        \item Each BPMN component represents BP's task. Tasks, then, are transformed into three queue system types: 
                    \begin{enumerate}[label=\roman*)]
                        \item Low skilled queues, $l_{y_i}$,
                        \item High skilled queues, $h_{y_j}$, and
                        \item Automated system queues, $x_{y_k}$.
                    \end{enumerate}
            
\noindent where $y$ can be identified either as $n$ or $a$ for non-automated and automated task respectively. Figure \ref{fig:BPTaskMapping} also shows the variable $\tau_y$ that represents the execution time of a $n$ or $a$ BP type.
                
    \end{itemize}
    
    To assess productivity variations, we defined a BP productivity variation ratio as:
        \begin{equation}
            \delta p := \dfrac{p_a}{p_n}
            \label{eq:BPdP}
        \end{equation}
\noindent where $p_n$ and $p_a$ are the labour productivity values of the BP before adopting the activity automation and after respectively.

    Adapting equation \ref{eq:eq1}, and formalising our initial statements, we obtained the following: 
                        \begin{equation}
                            \delta p = \dfrac{\tau_n}{\tau_a}
                            \label{eq:MiCtnta}
                        \end{equation}
                        \begin{equation}
                            \tau_a =
                               \begin{cases}
                                    (1 - \psi) \tau_n \text{,} &\text{if }\delta p \ge 1 \\
                                    (1 + \psi) \tau_n \text{,} &\text{if } 0 < \delta p < 1 \\
                               \end{cases}\quad \text{.}
                               \label{eq:MiCtau}
                        \end{equation}
            
\noindent where $\psi$ is the change rate in BP execution time, due to the automation, figure \ref{fig:BPPhiVal} describes this behaviour. 
    From \ref{eq:MiCtnta} and \ref{eq:MiCtau}:
                    \begin{equation}
                        \psi =
                            \begin{cases}
                                (1 - \dfrac{1}{\delta p}) \text{,} &\text{if }\tau_n \ge \tau_a \\
                                (\dfrac{1}{\delta p} - 1) \text{,} &\text{if }\tau_n < \tau_a \\
                            \end{cases}\quad \text{.}
                        \label{eq:Micdpresu}
                    \end{equation}

            We can dissect $\psi$ into two components: the contribution rate due to the localised automation gain, $\gamma$,and the contribution of the impact on the other tasks due to its integration, $\theta$ . See figure \ref{fig:BPPhiVal}.
            
    \noindent Therefore:    
            \begin{equation}
                        \psi = 
                            \begin{cases}
                                \theta + \gamma \text{,} &\text{if } \tau_n \ge \tau_a \\
                                \\
                            \theta - \gamma \text{,} &\text{if } \tau_n < \tau_a  \\
                            \end{cases}  \text{.}
                        \label{eq:MicIcomp} 
                    \end{equation} 
            Thus, we can estimate $\gamma$ in the following way: 
                    \begin{equation}
                        \gamma = \kappa \psi  \quad \text{.}
                        \label{eq:Micgamma} 
                    \end{equation}
            $\kappa$ represents the promised improvement in the execution time due to the introduction of the automated activity. In this fashion, $\gamma$ denotes the actual improvement rate.
            
            Considering the complete form of equation \ref{eq:BPdP}, we have the following:
            
         \begin{figure}
            \centering
                \begin{minipage}{.5\textwidth}
                    \centering
                        \includegraphics[width=.9\linewidth]{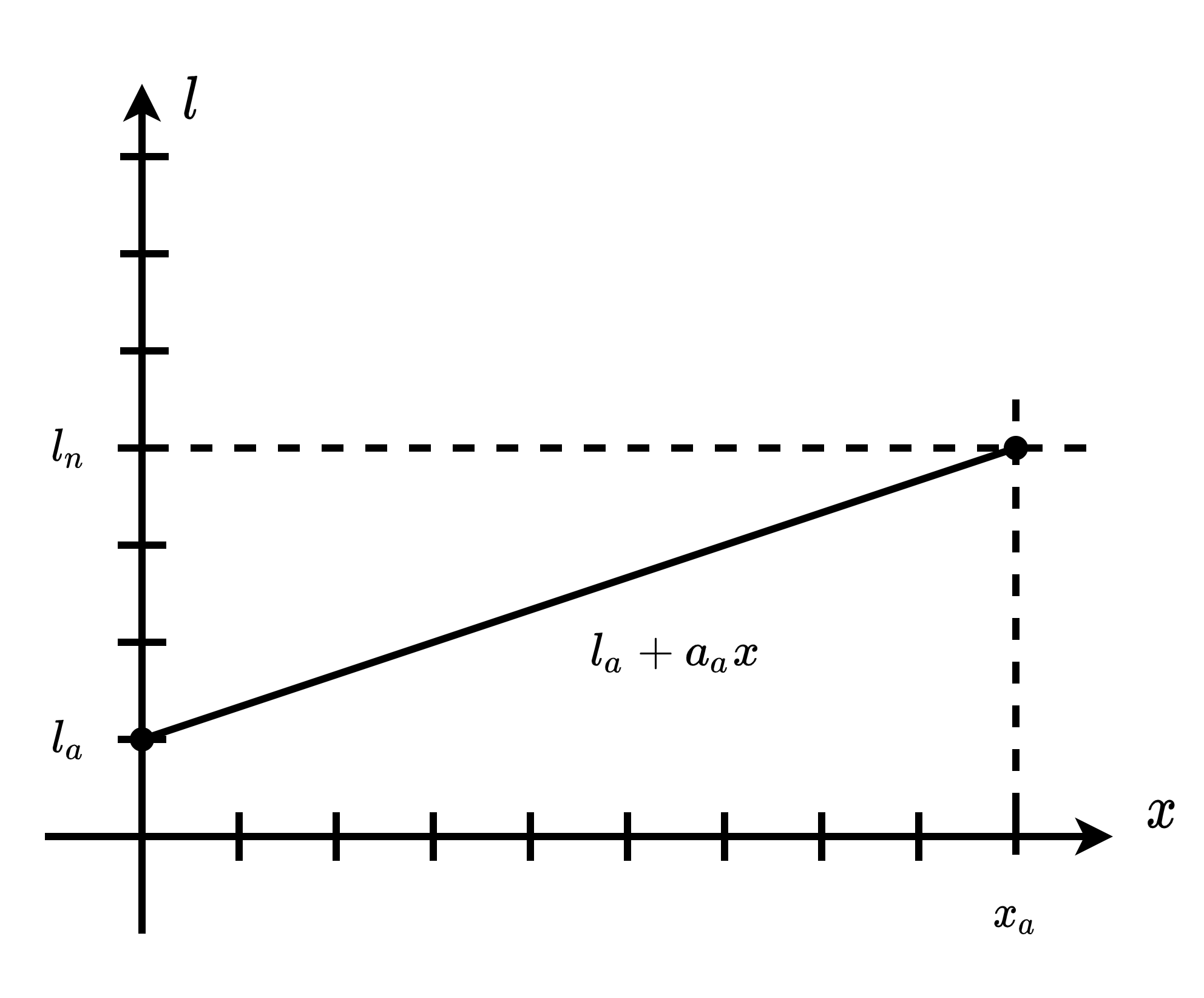}
                        \caption{Linear low-skill worker \\ distribution model.}
                        \label{fig:LLSKWDM}
                        \end{minipage}%
                \begin{minipage}{.5\textwidth}
                    \centering
                        \includegraphics[width=.9\linewidth]{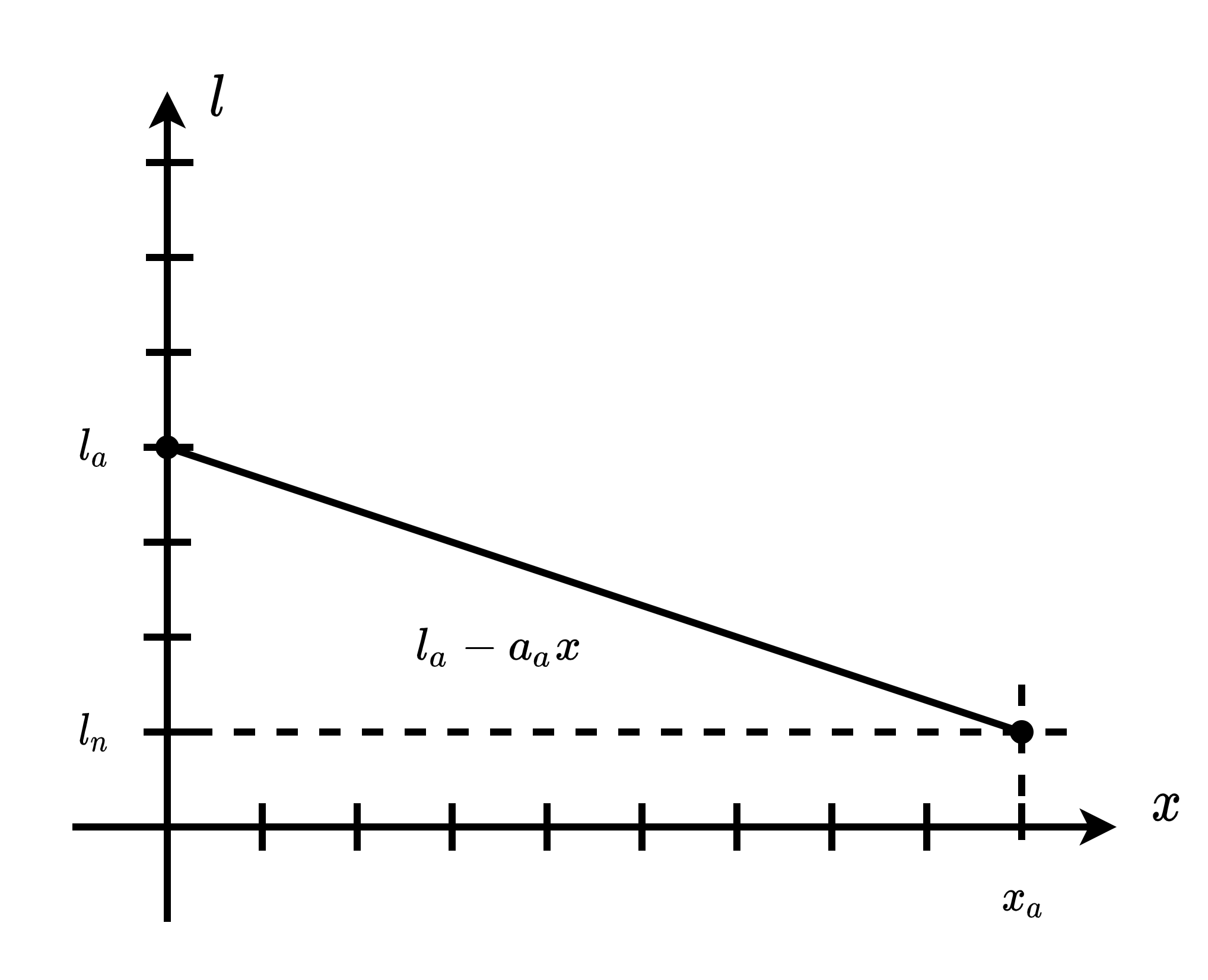}
                        \caption{Linear low-skill worker \\ distribution model negative slope.}
                        \label{fig:LLSKWDMN}
                \end{minipage}
        \end{figure}

        \begin{equation}
            \delta p =
                    \dfrac{\tau_n}{\tau_a} \Bigg ( \dfrac{l_a + a_a x_a}{l_n} \Bigg )^{\alpha} \Bigg ( \dfrac{h_a}{h_n}  \Bigg )^{(1 - \alpha)} 
                    \quad \text{.}
                    \label{eq:MiCDP}
        \end{equation}
            
            Figures \ref{fig:LLSKWDM} and \ref{fig:LLSKWDMN} show the geometric interpretation of the low-skilled ($ls$) worker distribution. The figure \ref{fig:LLSKWDM} shows the case when the introduction of automation reduces the number of low-skilled worker hours. $l_a$ should move towards zero whereas $x$ should grow. Thus $l_a+a_ax_a$tend to $l_n$ at a rate of $a_a$.
            The figure \ref{fig:LLSKWDMN} shows the case when the introduction of automation produces an increment in the number of low-skilled worker hours. These changes indicate that we would require more personnel to execute the process. In this case, the effect over $l_a$ surpasses the value of $l_n$. As in the previous case, $l_a+a_ax_a$ should reach the value of $l_n$ at a descendent rate of $a_a$.     
            
            Thus, grouping and simplifying terms, we have:    
                    \begin{subequations}
                        \begin{align}
                            h_a &= h_n  \quad \text{,} \\
                             R &= ( 1 - \dfrac{l_a}{l_n})  \quad \text{,} \\
                            a_a  &=
                                  \dfrac{l_n - l_a}{x_a}   \quad \text{.}
                        \end{align}  
                    \label{eq:aFactors} 
                    \end{subequations}

\noindent where $R$ is the share of hours at potential risk for becoming redundant due to the integration of an automated activity. 

   \section{Results}
    \subsection{Productivity performance and labour composition}
\begin{figure}[htbp]
                \centering 
                    {\includegraphics[scale=0.3]{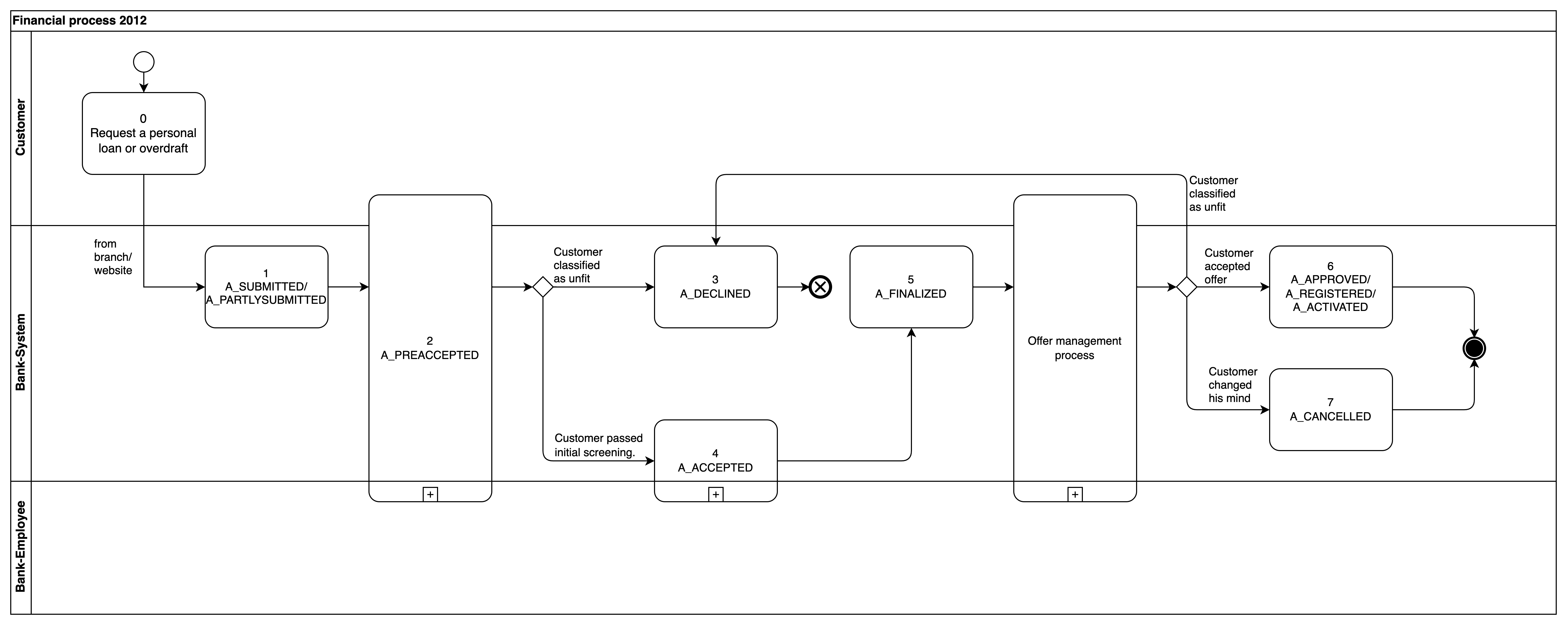}} 
                \caption{Business process 2012.}
                \label{fig:BP2012} 
            \end{figure}

\begin{figure}[htbp]
                \centering 
                    {\includegraphics[scale=0.06]{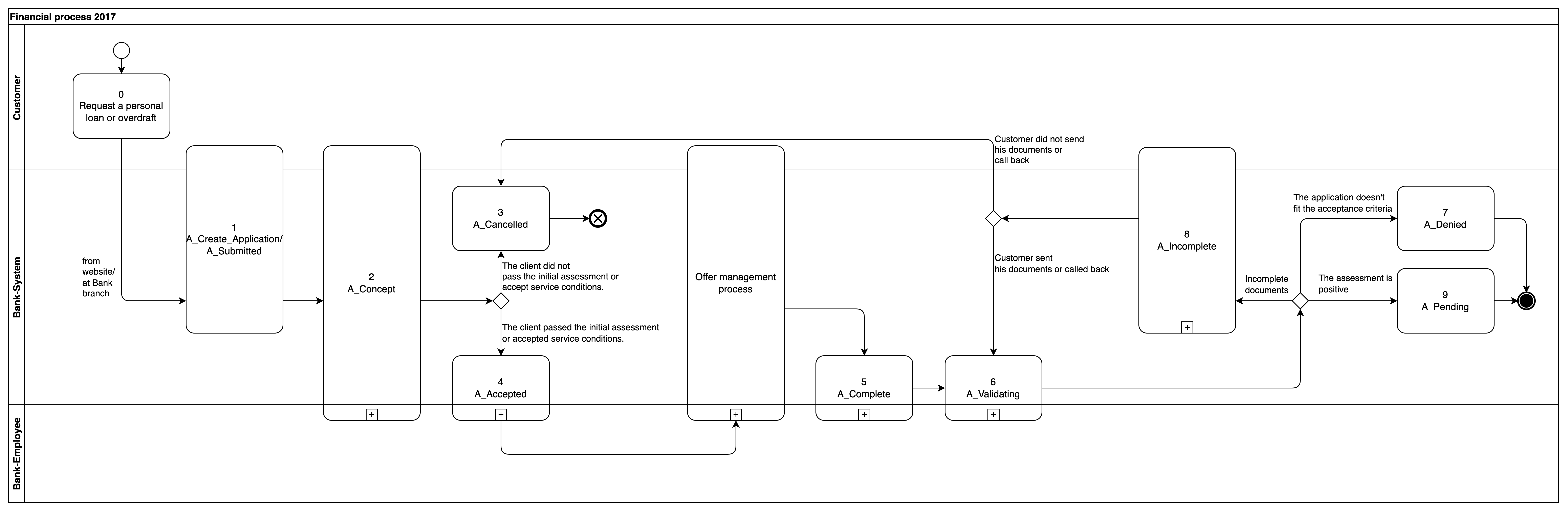}} 
                \caption{Business process 2017.}
                \label{fig:BP2017} 
            \end{figure}

        After reconstructing the BPs from both logs, figs \ref{fig:BP2012} and \ref{fig:BP2017}, we identified the main paths of each process. A main path is a series of activities that go from an initial state to a final state \cite{Kalnins2004}.
        
        We found three main paths. See table \ref{tab:MP1217}. Paths describe the initial and final process state. As we can see, these paths are equivalent in 2012 and 2017. Each path contains an automated task. The process state names changed between 2012 and 2017.
        
        Path A describes the scenario when the loan request did not accomplish a requirement. Path B represents the case when the financial institution approved the loan request and the customer accepted the authorised loan offer. Finally, Path C shows the situation when the customer neither accepted the loan offer, supplied the required documentation nor wanted the loan last minute.
        
                \begin{table}[htbp]
                    \begin{tabular}{|c|c|c|}
                     \hline
                        Path name & 2012 & 2017 \\ 
                        \hline 
                        A & $A\_SUBMITTED \rightarrow A\_CANCELLED$   & $A\_Create \quad Application \rightarrow  A\_Denied$ \\ \hline 
                        B & $A\_SUBMITTED \rightarrow A\_REGISTERED$   &
                        $A\_Create \quad Application \rightarrow A\_Pending$ \\ \hline 
                        C & $A\_SUBMITTED \rightarrow A\_DECLINED$   &  
                        $A\_Create \quad Application \rightarrow A\_Cancelled$ \\ \hline
                    \end{tabular}
                    \caption{Main paths 2012 and 2017}
                    \label{tab:MP1217}
                \end{table}
                    
        \begin{figure}
            \centering
                \begin{minipage}{.5\textwidth}
                    \centering
                        \includegraphics[width=.9\linewidth]{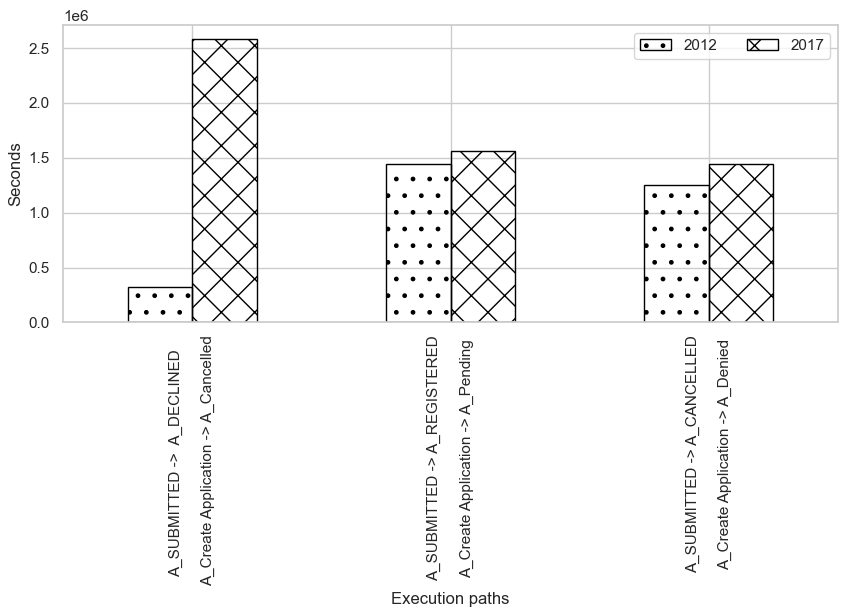}
                        \caption{Terminal paths - Average execution time.}
                        \label{fig:PPWCT}
                        \end{minipage}%
                \begin{minipage}{.5\textwidth}
                    \centering
                        \includegraphics[width=.9\linewidth]{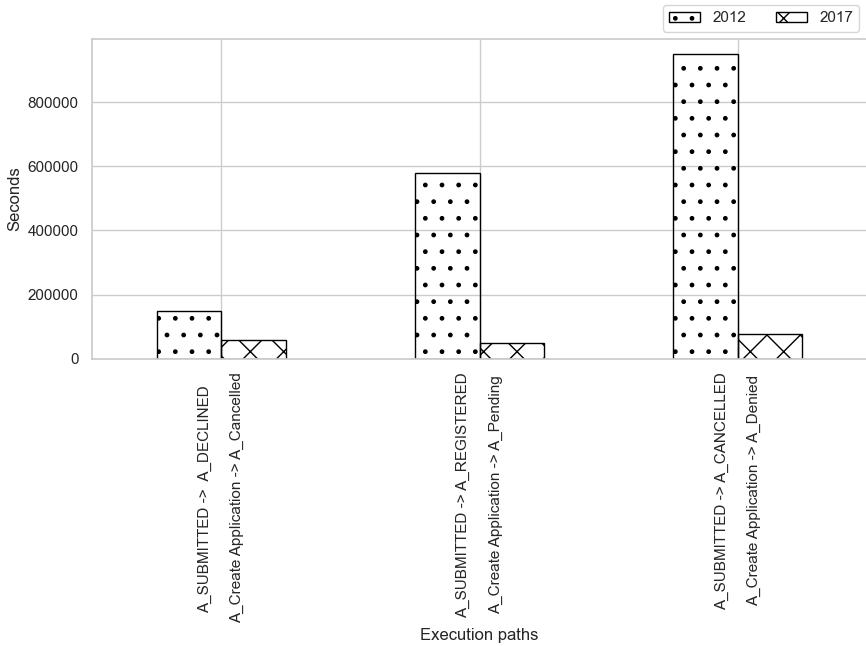}
                        \caption{Terminal paths - Average execution time without process customer time.}
                        \label{fig:PPWoCT}
                \end{minipage}
        \end{figure}
                    
       Then, we calculated the average process duration per path. See figures \ref{fig:PPWCT} and \ref{fig:PPWoCT}.
        
        Figure \ref{fig:PPWCT} shows the total execution time. These results contain all process activities. Even those that exclusively depend on the customers. 
        Customer activities introduce overhead in the process. This factor considerably impacts execution times and productivity.
        
        On the other hand, figure \ref{fig:PPWoCT} portrays the execution without customer activities.  Performance, in general, displays improvements.
        
        These two measured scenarios represented a unique opportunity to observe and characterise the impact of the automation intervention.
        In both cases, there is labour displacement. Tables \ref{tab:resUPM} and \ref{tab:resUPNM} show the productivity results, and \ref{tab:ldAS2ADE} to \ref{tab:ldACA2AD} show the labour displacement per path.

        \begin{table}[!htb]
            \begin{minipage}{.5\linewidth}
                    \centering
                    \begin{tabular}{l|c|c|c|}
                        \cline{2-4}
                            & A & B & C \\ \hline
                    \multicolumn{1}{|l|}{$\delta p$ } &0.12 & 0.92 & 0.87 \\ \hline
                    \multicolumn{1}{|l|}{$\kappa$}  &0.34 & 0.31	& 0.77  \\ \hline
                    \multicolumn{1}{|l|}{$\psi$}  &7.03& 0.09 & 0.15   \\ \hline
                    \multicolumn{1}{|l|}{$\gamma$}    &2.36 &	0.027& 0.12 \\ \hline
                    \multicolumn{1}{|l|}{$\theta$}    &9.39&	0.11& 0.27 \\ \hline
                \end{tabular}
                    \caption{Metrics by execution path with customer time.}
                   \label{tab:resUPM}
            \end{minipage}%
            \begin{minipage}{.5\linewidth}
                \centering
            \begin{tabular}{l|c|c|c|}
            \cline{2-4}
                                     & A & B & C \\ \hline
            \multicolumn{1}{|l|}{$\delta p$ }  &  2.47 & 11.70 & 12.56 \\ \hline
            \multicolumn{1}{|l|}{$\kappa$} &0.07 & 0.07 & 0.14 \\ \hline
            \multicolumn{1}{|l|}{$\psi$}  &0.60 & 0.91 &	0.92  \\ \hline
            \multicolumn{1}{|l|}{$\gamma$}    &	0.04  & 0.07 & 0.13  \\ \hline
            \multicolumn{1}{|l|}{$\theta$}    &0.55 & 0.85 &	0.79    \\ \hline
          \end{tabular}
          \caption{Metrics by execution path w/o customer time.}
          \label{tab:resUPNM}
            \end{minipage} 
        \end{table}
    
        \begin{table}[!htb]
            \begin{minipage}{.5\linewidth}
            \centering
                \begin{tabular}{|l|c|c|}
                    \cline{1-3}     Variable
                                    & $2012$ 
                                    & $2017$ \\  \hline
            \multicolumn{1}{|l|}{$H$ [s]} &\multicolumn{1}{|r|}{14481.39} & \multicolumn{1}{|r|}{189161.40}  \\ \hline
            \multicolumn{1}{|l|}{$L$ [s]}  &\multicolumn{1}{|r|}{70143.47} & \multicolumn{1}{|r|}{279719.92} \\ \hline
            \multicolumn{1}{|l|}{$X$ [s]}  & N/A & \multicolumn{1}{|r|}{272510.87} \\ \hline
            \multicolumn{1}{|l|}{$A$ [dimensionless]}  & N/A & \multicolumn{1}{|r|}{-0.77} \\ \hline
            \multicolumn{1}{|l|}{$R$ [dimensionless]}  & N/A &\multicolumn{1}{|r|}{-2.99} \\ \hline
          \end{tabular}
          \caption{Average path A labour composition with \\ customer time.}
          \label{tab:ldAS2ADE}
            \end{minipage}%
            \begin{minipage}{.5\linewidth}
            \centering
            \begin{tabular}{|l|c|c|}
                    \cline{1-3}     Variable
                                    & $2012$ 
                                    & $2017$ \\  \hline
                    \multicolumn{1}{|l|}{$H$ [s]} & \multicolumn{1}{|r|}{23431.80} & \multicolumn{1}{|r|}{8207.89}  \\ \hline
                    \multicolumn{1}{|l|}{$L$ [s]}  &\multicolumn{1}{|r|}{54256.01} & \multicolumn{1}{|r|}{206612.05} \\ \hline
                    \multicolumn{1}{|l|}{$X$ [s]}  & N/A & \multicolumn{1}{|r|}{8215.09} \\ \hline
                    \multicolumn{1}{|l|}{$A$ [dimensionless]}  & N/A & \multicolumn{1}{|r|}{-18.55} \\ \hline
                    \multicolumn{1}{|l|}{$R$ [dimensionless]}  & N/A & \multicolumn{1}{|r|}{-2.81} \\ \hline
            \end{tabular}
            \caption{Average path A labour composition w/o \\ customer time.}
            \label{tab:ldACA2AC}
            \end{minipage} 
        \end{table}

            \begin{table}[!htb]
            \begin{minipage}{.5\linewidth}
            \centering
                \begin{tabular}{|l|c|c|}
                    \cline{1-3}     Variable
                                    & $2012$ 
                                    & $2017$ \\  \hline
            \multicolumn{1}{|l|}{$H$ [s]} &\multicolumn{1}{|r|}{11586.73} & \multicolumn{1}{|r|}{51000.80}  \\ \hline
            \multicolumn{1}{|l|}{$L$ [s]}  &\multicolumn{1}{|r|}{78617.66} & \multicolumn{1}{|r|}{164189.45} \\ \hline
            \multicolumn{1}{|l|}{$X$ [s]}  & N/A & \multicolumn{1}{|r|}{114515.37} \\ \hline
            \multicolumn{1}{|l|}{$A$ [dimensionless]}  & N/A & \multicolumn{1}{|r|}{-0.75} \\ \hline
            \multicolumn{1}{|l|}{$R$ [dimensionless]}  & N/A &\multicolumn{1}{|r|}{-1.09} \\ \hline
          \end{tabular}
          \caption{Average path B labour composition with \\ customer time.}
          \label{tab:ldAS2ARE}
            \end{minipage}%
            \begin{minipage}{.5\linewidth}
            \centering
                \begin{tabular}{|l|c|c|}
                    \cline{1-3}     Variable
                                    & $2012$ 
                                    & $2017$ \\  \hline
            \multicolumn{1}{|l|}{$H$ [s]} &\multicolumn{1}{|r|}{27733.48} & \multicolumn{1}{|r|}{5457.20}
            \\ \hline
            \multicolumn{1}{|l|}{$L$ [s]}  &\multicolumn{1}{|r|}{49796.97} & \multicolumn{1}{|r|}{223307.38} \\ \hline
            \multicolumn{1}{|l|}{$X$ [s]}  & N/A & \multicolumn{1}{|r|}{36512.86} \\ \hline
            \multicolumn{1}{|l|}{$A$ [dimensionless]}  & N/A & \multicolumn{1}{|r|}{-4.75} \\ \hline
            \multicolumn{1}{|l|}{$R$ [dimensionless]}  & N/A &\multicolumn{1}{|r|}{-3.48} \\ \hline
          \end{tabular}
          \caption{Average path B labour composition w/o \\ customer time.}
          \label{tab:ldACA2AP}
            \end{minipage} 
        \end{table}
    
                \begin{table}[!htb]
            \begin{minipage}{.5\linewidth}
            \centering
                \begin{tabular}{|l|c|c|}
                    \cline{1-3}     Variable
                                    & $2012$ 
                                    & $2017$ \\  \hline
            \multicolumn{1}{|l|}{$H$ [s]} &\multicolumn{1}{|r|}{84905.26} & \multicolumn{1}{|r|}{88975.25}  \\ \hline
            \multicolumn{1}{|l|}{$L$ [s]}  &\multicolumn{1}{|r|}{62464.23} & \multicolumn{1}{|r|}{179400.26} \\ \hline
            \multicolumn{1}{|l|}{$X$ [s]}  & N/A & \multicolumn{1}{|r|}{121752.20} \\ \hline
            \multicolumn{1}{|l|}{$A$ [dimensionless]}  & N/A & \multicolumn{1}{|r|}{-0.96} \\ \hline
            \multicolumn{1}{|l|}{$R$ [dimensionless]}  & N/A &\multicolumn{1}{|r|}{-1.87} \\ \hline
          \end{tabular}
          \caption{Average path C labour composition with \\ customer time.}
          \label{tab:ldAS2ACA}
            \end{minipage}%
            \begin{minipage}{.5\linewidth}
            \centering
                \begin{tabular}{|l|c|c|}
                    \cline{1-3}     Variable
                                    & $2012$ 
                                    & $2017$ \\  \hline
            \multicolumn{1}{|l|}{$H$ [s]} &\multicolumn{1}{|r|}{203499.39} & \multicolumn{1}{|r|}{328146.74}  \\ \hline
            \multicolumn{1}{|l|}{$L$ [s]}  &\multicolumn{1}{|r|}{93235.49} & \multicolumn{1}{|r|}{267421.80} \\ \hline
            \multicolumn{1}{|l|}{$X$ [s]}  & N/A & \multicolumn{1}{|r|}{33373.55} \\ \hline
            \multicolumn{1}{|l|}{$A$ [dimensionless]}  & N/A & \multicolumn{1}{|r|}{-5.22} \\ \hline
            \multicolumn{1}{|l|}{$R$ [dimensionless]}  & N/A &\multicolumn{1}{|r|}{-1.87} \\ \hline
          \end{tabular}
          \caption{Average path C labour composition w/o \\ customer time.}
          \label{tab:ldACA2AD}
            \end{minipage} 
        \end{table}

    \subsection{Queue system parameters}
   
    Tables \ref{tab:GQPWCT} and \ref{tab:GQPWoCT} show the recommended parameters to simulate under and over-performed processes. Queue parameters are presented as factors. Therefore, we should only multiply the current queue values and substitute them in our model to simulate the introduction of automation in a specific BP. We labelled the columns with the legend $[\times$  $n]$, which means: the factor times the value of the BP's non-automated queue.
    
        \begin{table}[htb]
            \begin{minipage}{.45\linewidth}
                    \centering
                        \begin{tabular}{|c|c|c|}
                        \cline{1-3}     Queue
                                        & $\lambda$  
                                        & $\mu$ \\
                                        type
                                        & [$\times$ n] 
                                        & [$\times$ n] \\ \hline
                        \multicolumn{1}{|l|}{$l$} &\multicolumn{1}{|r|}{2.29} & \multicolumn{1}{|r|}{0.04} \\ \hline
                        \multicolumn{1}{|l|}{$h$} &\multicolumn{1}{|r|}{2.05} & \multicolumn{1}{|r|}{\num{2.11e-5}} \\ \hline
                        \end{tabular}
                    \caption{Queue parameters for: \\ $n \rightarrow a$ with customer time.}
                    \label{tab:GQPWCT}
           \end{minipage}%
            \begin{minipage}{.45\linewidth}
                \centering
                        \begin{tabular}{|c|c|c|}
                        \cline{1-3}     Queue
                                        & $\lambda$  
                                        & $\mu$ \\
                                        type
                                        & [$\times$ n] 
                                        & [$\times$ n] \\ \hline
                        \multicolumn{1}{|l|}{$l$} &\multicolumn{1}{|r|}{2.29} & \multicolumn{1}{|r|}{0.41} \\ \hline
                        \multicolumn{1}{|l|}{$h$} &\multicolumn{1}{|r|}{2.05} & \multicolumn{1}{|r|}{0.02} \\ \hline
                        \end{tabular}
                    \caption{Queue parameters for: \\ $n \rightarrow a$  w/o customer time.}
                    \label{tab:GQPWoCT}
            \end{minipage} 
        \end{table}

         \begin{table}[htb]
            \begin{minipage}{.45\linewidth}
                    \centering
                        \begin{tabular}{|c|c|c|}
                        \cline{1-3}     Queue
                                        & $\lambda$  
                                        & $\mu$ \\
                                        type
                                        & [$customer/s$] 
                                        & [$customer/s$] \\ \hline
                        \multicolumn{1}{|l|}{$a$} & \multicolumn{1}{|r|}{1.10} & \multicolumn{1}{|r|}{1.29} \\ \hline
                        \end{tabular}
                    \caption{Queue parameters for: \\ $n \rightarrow a$ with customer time.}
                    \label{tab:GQAPWCT}
           \end{minipage}%
            \begin{minipage}{.45\linewidth}
                \centering
                        \begin{tabular}{|c|c|c|}
                        \cline{1-3}     Queue
                                        & $\lambda$  
                                        & $\mu$ \\
                                        type
                                        & [$customer/s$] 
                                        & [$customer/s$] \\ \hline
                        \multicolumn{1}{|l|}{$a$} & \multicolumn{1}{|r|}{1.10} & \multicolumn{1}{|r|}{6.01} \\ \hline
                        \end{tabular}
                    \caption{Queue parameters for: \\ $n \rightarrow a$  w/o customer time.}
                    \label{tab:GQAPWoCT}
            \end{minipage} 
        \end{table}
   
   As for the automated systems queues, table \ref{tab:GQAPWCT} and \ref{tab:GQAPWoCT} present the suggested parameters based on our results. Therefore, they are a starting point for the simulation of a BP automation intervention. 

\section{Discussion}
   
    Figures \ref{fig:PPWCT}  and \ref{fig:PPWoCT} show the execution times of the identified paths. We ordered the courses in pairs. The first name corresponds to the 2012 process and the second one to 2017. 
    Figure \ref{fig:PPWCT} exhibits that after the introduction of automation, the BP increased its average completion time. Table \ref{tab:resUPM} expands this information. Each column describes performance values per path. 
    As for path A, in table \ref{tab:resUPM}, we can observe that $\delta p$ shows a productivity decrement. $\kappa$ reveals that BP execution improved by 33.59\%. However, the counter effects represented by $\theta$ are meaningful; about 939.38\%. $\kappa$ and $\theta$ values mean that average time enhanced and decreased, respectively, in that proportion. Paths B and C display similar results.
    On the other hand, figure \ref{fig:PPWoCT} presents the opposite results. This chart did not consider tasks that depended on customer inputs. In other words, activities that are not accountable to the organisation. 
    Figure \ref{fig:PPWoCT} on the other hand shows the opposite outcomes. Task filtering is the justification for this behaviour. The jobs that were solely liable to the BP client were filtered out. For instance, 'Request form completion' or 'Loan acceptance'. As a result, figure \ref{fig:PPWoCT} only shows the duration of the activities where the agent is an employee of the company.
    Table \ref{tab:resUPNM} shows that the evaluation of non-customer-based tasks improved the productivity variations indicator $\delta p$. $\kappa$ and $\theta$ have values smaller than one. This feature is a sign that the change in the BP improved productivity. $\delta p$, for all the paths, shows considerable improvements.  We can infer,  then, that customer-dependent tasks may introduce distortions to productivity evaluations. The misshaping consists of including the BP client execution time. In this manner, one request may appear to take longer to process than cases where only the firm's personnel interveined. As a result, averaged metrics may produce results that do not correspond to BP performance.
    We can also notice that $\delta p$ values larger than one means that the adoption of the new technology in the analysed BP improved labour productivity.
    As for the labour displacement, tables \ref{tab:ldAS2ADE} to \ref{tab:ldACA2AD} show the composition per path. Each table contains the value of the average execution of high-skilled, low-skilled, and automated system hours.
    They also include $A$ and $R$ values. Our model considers that all values should be positive. In the case of $A$, a positive number indicates a surge in the number of automated hours until they replace the intervention of low-skilled workers. A negative value for $R$ denotes that we need extra low-skilled worker hours to complete the task. Thus, we can infer that labour costs may increase after the introduction of automation.
    Furthermore, $A$ describes how quickly technological advancements would reduce low-skilled worker hours. We anticipate a $L$ reduction in this proportion whenever any other BP task uses the same technology. However, different technologies may change this value.

    Another interesting finding is the $R$ value for path B. See tables 2 and 6. The majority of the loan requests are grouped in Path B. Despite having the best productivity variation, $R$ shows that the BP would need to hire more low-skilled workers.
    We also discovered that after removing customer-dependent activities, the process would require roughly twice as many $L$ low-skilled hours as the initial scenario. See the "2017" column in tables 6 and 7.

    Process reports can explain this behaviour. Automation introduced the possibility of processing a higher number of loan requests. Each request requires the customer to provide additional information and documents. Thus, the institution requires that bank employees track the request and ask the customer to complete the process. Therefore, more requests imply more agents.
    Both scenarios reflect this feature. We can infer that labour composition and productivity variations are not related. Moreover, productivity improvement does not directly imply less low-skilled worker hours. The introduction of automation may cause an increment in this variable as well.

    Tables \ref{tab:GQPWCT} and \ref{tab:GQPWoCT} show that with or without response customer time, the waiting time duplicated its value; as for the service time, it reduced the pre-automation service rate. We presented these weights as factors to simulate the integration of the automation intervention in a BP. Therefore, process designers only have to multiply their current queues system values by the proposed rates to obtain the recommended simulation values.

    On the other hand, tables \ref{tab:GQAPWCT} and \ref{tab:GQAPWoCT}, automated systems' queues, contain the parameters we found in our test case. These values show that waiting time remained the same and service time raised. This phenomenon may appear contradictory; we would expect increments in the other queues' service time parameters. 
    AI queues seem to accelerate the process. Their quick output means more considerable waiting times for the rest of the BP, and a labour redistribution implies an adjustment in service times. 

    In our test case, we required more low-skilled hours to complete a considerable backlog generated by the integration of automation into the BP. We can infer that our model is capturing the BP's natural behaviour. 
    

\section{Conclusions}
   
   Our outcomes show that it is possible to calculate productivity variations with only BP execution times. As a result, the analysis can be scaled down to the task level (RQ1). Our experiment divided the review into three sections. Each of them examines an execution path. We observed that introducing automation did not affect productivity. After that, it decreased. That is, Solow's claim was proven correct. However, a closer examination of the firm's solely responsible tasks revealed that productivity increased significantly. The explanatory argument of productivity mismeasurement is then supported. In other words, standard evaluation methods equalise benefits, producing misleading results. Thus, we provided empirical evidence to address Solow's automation paradox in total.
    Our model simplified labour productivity calculation (RQ2). Based on microeconomic theory and on the feasibility of process mining data collection, we analysed a BP before and after the introduction of an automation intervention system, all mediated by a parametrised queue system, in close dialogue with the Cobb-Douglas abstraction of labor distribution. BP workflows exhibit a granular representation to identify the affected activity paths (RQ3). At this level, it is possible to isolate those routes that are not influenced by the automation of some tasks. Thus, irrelevant components do not affect productivity calculation (RQ4).  
    
     Furthermore, the introduction of a production function that shifted the focus from cost to labour composition allows the analysis of the labour displacement. Consequently, we acquired information on the new labour distribution (R05).  

\section*{Acknowledgments}
This work was supported by the Science and Technology Council of Mexico (Consejo Nacional de Ciencia y Tecnología, CONACYT, https://www.conacyt.gob.mx).

\bibliographystyle{unsrt}  
\bibliography{references}

\begin{thebibliography}{10}

\bibitem{jalowiecki2020big}
Piotr Ja{\l}owiecki.
\newblock {Big Data Analysis for Management from Solow's Paradox Perspective in
  Polish Industry}.
\newblock In {\em Management in the Era of Big Data}, pages 177--192. Auerbach
  Publications, 2020.

\bibitem{Solow1957}
Robert~M. Solow.
\newblock {Technical Change and the Aggregate Production Function}.
\newblock {\em The Review of Economics and Statistics}, 39(3):312, aug 1957.

\bibitem{10.5555/171750.171780}
Gordon~B. Davis, Rosann~Webb Collins, Michael~A. Eierman, and William~D. Nance.
\newblock {\em Productivity from Information Technology Investment in Knowledge
  Work}, page 327–343.
\newblock IGI Global, USA, 1993.

\bibitem{ByRobertM.Solow1987WBWO}
Robert~M Solow.
\newblock {We'd Better Watch Out: MANUFACTURING MATTERS The Myth of the
  Post-Industrial Economy. By Stephen S. Cohen and John Zysman.}
\newblock {\em New York Times (1923-)}, page 297, 1987.

\bibitem{Brynjolfsson2017}
Erik Brynjolfsson, Daniel Rock, and Chad Syverson.
\newblock {Artificial Intelligence and the Modern Productivity Paradox: A Clash
  of Expectations and Statistics}.
\newblock Technical report, National Bureau of Economic Research, Cambridge,
  MA, nov 2017.

\bibitem{Brynjolfsson1996}
Erik Brynjolfsson and Lorin Hitt.
\newblock {Paradox Lost? Firm-Level Evidence on the Returns to Information
  Systems Spending}.
\newblock {\em Management Science}, 42(4):541--558, jul 1996.

\bibitem{Dongen2020}
Boudewijn van Dongen.
\newblock {BPI Challenges - IEEE Task Force on Process Mining}, 2020.

\bibitem{Bonsay_Cruz_Firozi_Camaro_2021}
Jamielyn~O. Bonsay, Abigail~P. Cruz, Homa~C. Firozi, and Peter Jeff~C. Camaro.
\newblock Artificial intelligence and labor productivity paradox: The economic
  impact of ai in china, india, japan, and singapore.
\newblock {\em Journal of Economics, Finance and Accounting Studies},
  3(2):120–139, Nov. 2021.

\bibitem{Denning2021}
Steve Denning.
\newblock {Why Computers Didn't Improve Productivity}, 2021.

\bibitem{Schweikl2020}
Stefan Schweikl and Robert Obermaier.
\newblock {Lessons from three decades of IT productivity research: towards a
  better understanding of IT-induced productivity effects}.
\newblock {\em Management Review Quarterly}, 70(4):461--507, nov 2020.

\bibitem{63379290255c45f3b8ceb002d050138d}
Joel Mokyr.
\newblock Secular stagnation? not in your life.
\newblock {\em Geneva Reports on the World Economy}, 83(August 2014):83--89,
  January 2014.

\bibitem{back2022return}
Asta B{\"{a}}ck, Arash Hajikhani, Angela J{\"{a}}ger, Torben Schubert, Arho
  Suominen, and Others.
\newblock {Return of the Solow-paradox in AI? AI-adoption and firm
  productivity}.
\newblock Technical report, Lund University, CIRCLE-Centre for Innovation
  Research, 2022.

\bibitem{Acemoglu2014}
Daron Acemoglu, David Autor, David Dorn, Gordon~H. Hanson, and Brendan Price.
\newblock {Return of the Solow Paradox? IT, Productivity, and Employment in US
  Manufacturing}.
\newblock {\em American Economic Review}, 104(5):394--399, may 2014.

\bibitem{Du2022}
Longzheng Du and Weifen Lin.
\newblock {Does the application of industrial robots overcome the Solow
  paradox? Evidence from China}.
\newblock {\em Technology in Society}, 68:101932, feb 2022.

\bibitem{Acemoglu2018}
Daron Acemoglu and Pascual Restrepo.
\newblock {The Race between Man and Machine: Implications of Technology for
  Growth, Factor Shares, and Employment}.
\newblock {\em American Economic Review}, 108(6):1488--1542, jun 2018.

\bibitem{Jackson2019}
Matthew~O. Jackson and Zafer Kanik.
\newblock {How Automation that Substitutes for Labor Affects Production
  Networks, Growth, and Income Inequality}.
\newblock {\em SSRN Electronic Journal}, 2019.

\bibitem{Li2022}
Mengmeng Li and Xiaochuan Guo.
\newblock {Research on the Productivity Paradox of Information Technology}.
\newblock {\em Proceedings of Business and Economic Studies}, 5(2):19--27, apr
  2022.

\bibitem{Lu2021}
Yingying Lu and Yixiao Zhou.
\newblock {A review on the economics of artificial intelligence}.
\newblock {\em Journal of Economic Surveys}, 35(4):1045--1072, sep 2021.

\bibitem{Jones2010}
Charles~I. Jones and Paul~M. Romer.
\newblock {The New Kaldor Facts: Ideas, Institutions, Population, and Human
  Capital}.
\newblock {\em American Economic Journal: Macroeconomics}, 2(1):224--245, jan
  2010.

\bibitem{brynjolfsson2018artificial}
Erik Brynjolfsson, Daniel Rock, and Chad Syverson.
\newblock {Artificial intelligence and the modern productivity paradox: A clash
  of expectations and statistics}.
\newblock In {\em The economics of artificial intelligence: An agenda}, pages
  23--57. University of Chicago Press, 2018.

\bibitem{Arnold2013}
G.~Arnold.
\newblock {\em {Essentials of corporate financial management}}.
\newblock Pearson, second edi edition, 2013.

\bibitem{Acemoglu2018b}
Daron Acemoglu and Pascual Restrepo.
\newblock {Artificial Intelligence, Automation and Work}.
\newblock Technical report, National Bureau of Economic Research, Cambridge,
  MA, jan 2018.

\bibitem{Eggertsson2019}
Gauti~B. Eggertsson, Neil~R. Mehrotra, and Jacob~A. Robbins.
\newblock {A Model of Secular Stagnation: Theory and Quantitative Evaluation}.
\newblock {\em American Economic Journal: Macroeconomics}, 11(1):1--48, jan
  2019.

\bibitem{VanderAalst2012}
Wil van~der Aalst.
\newblock {Process Mining}.
\newblock {\em ACM Transactions on Management Information Systems}, 3(2):1--17,
  jul 2012.

\bibitem{Diba2020}
Kiarash Diba, Kimon Batoulis, Matthias Weidlich, and Mathias Weske.
\newblock {Extraction, correlation, and abstraction of event data for process
  mining}.
\newblock {\em WIREs Data Mining and Knowledge Discovery}, 10(3), may 2020.

\bibitem{Garcia2019}
Cleiton dos~Santos Garcia, Alex Meincheim, Elio~Ribeiro {Faria Junior},
  Marcelo~Rosano Dallagassa, Denise Maria~Vecino Sato, Deborah~Ribeiro
  Carvalho, Eduardo Alves~Portela Santos, and Edson~Emilio Scalabrin.
\newblock {Process mining techniques and applications – A systematic mapping
  study}.
\newblock {\em Expert Systems with Applications}, 133:260--295, nov 2019.

\bibitem{Munoz-Gama2022}
Jorge Munoz-Gama, Niels Martin, Carlos Fernandez-Llatas, Owen~A. Johnson,
  Marcos Sep{\'{u}}lveda, Emmanuel Helm, Victor Galvez-Yanjari, Eric Rojas,
  Antonio Martinez-Millana, Davide Aloini, Ilaria~Angela Amantea, Robert
  Andrews, Michael Arias, Iris Beerepoot, Elisabetta Benevento, Andrea
  Burattin, Daniel Capurro, Josep Carmona, Marco Comuzzi, Benjamin Dalmas, Rene
  de~la Fuente, Chiara {Di Francescomarino}, Claudio {Di Ciccio}, Roberto
  Gatta, Chiara Ghidini, Fernanda Gonzalez-Lopez, Gema Ibanez-Sanchez, Hilda~B.
  Klasky, Angelina {Prima Kurniati}, Xixi Lu, Felix Mannhardt, Ronny Mans, Mar
  Marcos, Renata {Medeiros de Carvalho}, Marco Pegoraro, Simon~K. Poon, Luise
  Pufahl, Hajo~A. Reijers, Simon Remy, Stefanie Rinderle-Ma, Lucia Sacchi,
  Fernando Seoane, Minseok Song, Alessandro Stefanini, Emilio Sulis,
  Arthur~H.M. ter Hofstede, Pieter~J. Toussaint, Vicente Traver, Zoe
  Valero-Ramon, Inge van~de Weerd, Wil~M.P. van~der Aalst, Rob Vanwersch,
  Mathias Weske, Moe~Thandar Wynn, and Francesca Zerbato.
\newblock {Process mining for healthcare: Characteristics and challenges}.
\newblock {\em Journal of Biomedical Informatics}, 127:103994, mar 2022.

\bibitem{Jacobo-Romero2021}
Mauricio Jacobo-Romero and Andr{\'{e}} Freitas.
\newblock {Microeconomic foundations of decentralised organisations}.
\newblock In {\em Proceedings of the 36th Annual ACM Symposium on Applied
  Computing}, pages 282--290, New York, NY, USA, mar 2021. ACM.

\bibitem{Macdonald2000}
Stuart Macdonald, Pat Anderson, and Dieter Kimbel.
\newblock {Measurement or Management?: Revisiting the Productivity Paradox of
  Information Technology}.
\newblock {\em Vierteljahrshefte zur Wirtschaftsforschung}, 69(4):601--617, oct
  2000.

\bibitem{Atrill2017}
Peter Atrill and Eddie McLaney.
\newblock {\em {Accounting and Finance for Non-Specialists}}.
\newblock Pearson Education Limited, 10 edition, 2017.

\bibitem{Lima2012}
Rui~M. Lima.
\newblock {Integrating Production Planning and Control Business Processes}.
\newblock {\em International Journal of Productivity Management and Assessment
  Technologies}, 1(4):1--21, oct 2012.

\bibitem{Kossak2016}
Felix Kossak, Christa Illibauer, Verena Geist, Christine Natschl{\"{a}}ger,
  Thomas Ziebermayr, Bernhard Freudenthaler, Theodorich Kopetzky, and
  Klaus-Dieter Schewe.
\newblock {\em {Hagenberg Business Process Modelling Method}}.
\newblock Springer International Publishing, Cham, 2016.

\bibitem{Kalnins2004}
Audris Kalnins.
\newblock {Business Modelling. Languages and Tools}.
\newblock In {\em Progress in Industrial Mathematics at ECMI 2002}, pages
  41--52. Springer Berlin Heidelberg, Berlin, Heidelberg, 2004.

\bibitem{Prokopenko1987}
Joseph Prokopenko.
\newblock {\em {Productivity Management: A Practical Handbook}}.
\newblock International Labour Office, Geneva, 1987.

\bibitem{Brynjolfsson1996B}
Erik Brynjolfsson and Shinkyu Yang.
\newblock Information technology and productivity: A review of the literature.
\newblock In Marvin~V. Zelkowitz, editor, {\em Advances in Computers},
  volume~43 of {\em Advances in Computers}, pages 179--214. Elsevier, 1996.

\bibitem{Ross97}
Sheldon~M. Ross.
\newblock {\em Introduction to Probability Models}.
\newblock Academic Press, San Diego, CA, USA, sixth edition, 1997.

\bibitem{vanDongen2012}
Boudewijn van Dongen.
\newblock Bpi challenge 2012, 4 2012.

\bibitem{vanDongen2017}
Boudewijn van Dongen.
\newblock Bpi challenge 2017, 2 2017.

\bibitem{Heathfield1971}
David~F. Heathfield.
\newblock {The Cobb-Douglas Function}.
\newblock In {\em Production Functions}, pages 29--44. Macmillan Education UK,
  London, 1971.

\end{thebibliography}

\end{document}